%% file: CameraReady2027.tex
\documentclass[letterpaper]{article} 
\usepackage{aaai2027}  
\usepackage[hyphens]{url}  
\usepackage{graphicx} 
\usepackage{natbib}  
\usepackage{caption} 
\usepackage{algorithm}
\usepackage{algorithmic}

\usepackage{amsmath}
\usepackage{amssymb}

\usepackage{newfloat}
\usepackage{listings}
\DeclareCaptionStyle{ruled}{labelfont=normalfont,labelsep=colon,strut=off} 
\floatstyle{ruled}
\newfloat{listing}{tb}{lst}{}
\floatname{listing}{Listing}

\usepackage{booktabs}
\usepackage{subcaption}

\usepackage{multirow}
\usepackage{graphicx}
\usepackage{colortbl}
\usepackage{xcolor}

\definecolor{groupblue}{RGB}{235,240,248}
\definecolor{oursgreen}{RGB}{232,245,233}
\definecolor{ablationyellow}{RGB}{255,248,220}

\newcommand{\methodgroup}[1]{
\addlinespace[5pt]
\rowcolor{groupblue}
\textcolor{black!70}{\textbf{\textit{#1}}}
& & & & & & & & & & & & & & & & \\
\addlinespace[1pt]
}

\newcommand{\oursgroup}[1]{
\addlinespace[5pt]
\rowcolor{oursgreen}
\textbf{#1}
& & & & & & & & & & & & & & & & \\
\addlinespace[1pt]
}

\newcommand{\ablationgroup}[1]{
\addlinespace[5pt]
\rowcolor{ablationyellow}
\textcolor{black!70}{\textbf{\textit{#1}}}
& & & & & & & & & & & & & & & & \\
\addlinespace[1pt]
}

\title{Efficient Multimodal Generative Recommendation with\\ Latent Narrative Reasoning}
\author{
    Chenxing Wang\textsuperscript{\rm 1}\equalcontrib, Nantao Zheng\textsuperscript{\rm 1}\equalcontrib, Hao Miao\textsuperscript{\rm 2}, Juyuan Wang\textsuperscript{\rm 1}, Xinke Jiang, \\Yuchen Fang\textsuperscript{\rm 1}\corresponding, Aolin Li\textsuperscript{\rm 1}, Haijun Wu\textsuperscript{\rm 1}\\
}
\affiliations{
    \textsuperscript{\rm 1}Weixin Group, Tencent\\
    \textsuperscript{\rm 2}The Hong Kong Polytechnic University\\
    
    magnuswang@tencent.com, tmaczheng@tencent.com, hao-miao@outlook.com, jovwang@tencent.com, thinkerjiang@foxmail.com, fyclmiss@gmail.com, churenli@tencent.com, ethanhjwu@tencent.com
}

\begin{document}

\maketitle

\begin{abstract}
Generative recommendation reformulates item prediction as semantic
identifier generation, yet episodic content introduces a fundamentally
different setting where the target is determined by narrative evolution
rather than user preference. This task requires models to understand
multimodal storyline progression while addressing the efficiency
challenges caused by redundant visual contexts and costly explicit
reasoning generation. We propose \textbf{NarraLite}, an efficient
multimodal generative recommendation framework that jointly compresses
perception and reasoning. Specifically, Progressive Spectral
Compression selectively distills long visual contexts into compact
narrative-relevant evidence, preserving transition-critical information
while reducing redundant visual computation. Latent Narrative
Reasoning introduces context-routed latent reasoning tokens and aligns
their contextualized representations with future continuation semantics,
enabling implicit narrative inference without autoregressively decoding
textual rationales. We further establish a user-agnostic multimodal
benchmark for short-form drama continuation across UGC, PGC, and
OOD settings. Extensive experiments demonstrate that
NarraLite consistently improves continuation accuracy, narrative
coherence, and robustness over existing approaches, while achieving a
favorable accuracy--efficiency trade-off.
\end{abstract}


\input{secs/intro}

\input{secs/related}

\input{secs/method}

\input{secs/exp}

\input{secs/conclu}

\bibliography{aaai2027}


\end{document}

%% file: secs/intro.tex
\section{Introduction}

Generative recommendation reformulates item prediction as sequence
generation by representing items with discrete semantic identifiers
(SIDs) and directly generating target identifiers~\citep{li2024large,wang2024learnable}. However,
existing methods mainly focus on user preference modeling, while
episodic content requires predicting the segment that naturally follows
an ongoing storyline. As shown in Figure~\ref{fig:intro}, traditional
recommendation retrieves content based on user interests~\citep{chen2025onesearch,deng2025onerec}, whereas
narrative-aware recommendation requires understanding intrinsic event
transitions and predicting semantically and temporally coherent
continuations. This task further demands joint modeling of visual plot
evidence and textual descriptions, as either modality alone may miss
critical narrative cues.

Applying multimodal generative models to this setting, however, faces
two critical efficiency challenges~\citep{zhang2025beyond,cao2026fastdrivevla,he2024multi,zhan2026l2v}. First, long visual contexts contain
substantial redundancy: consecutive frames often repeat similar scenes,
characters, and backgrounds, while only a small subset of events is
essential for determining future storyline transitions. Processing all
visual tokens therefore introduces considerable computational overhead,
whereas aggressive compression may remove subtle but decisive narrative
cues. Second, explicit chain-of-thought (CoT) reasoning~\citep{yue2025cot4rec} introduces
additional autoregressive decoding costs by requiring the model to
generate lengthy intermediate rationales before predicting the final
SID. Although such reasoning can improve structured inference, the
generated explanations are unnecessary for recommendation and increase
latency. Efficient narrative recommendation therefore requires retaining
transition-critical visual evidence while enabling structured reasoning
without explicitly generating reasoning chains.

\input{figures/para}

To address these challenges, we propose \textbf{NarraLite}, an efficient
multimodal generative recommendation framework that jointly compresses
perception and reasoning. NarraLite introduces two complementary
components. \textbf{Progressive Spectral Compression (PSC)} selectively
distills long visual contexts into compact narrative-relevant tokens.
Instead of uniformly processing all visual observations, PSC identifies
visual evidence that provides complementary information to textual plot
descriptions, reducing redundant multimodal computation while preserving
important storyline cues. \textbf{Latent Narrative Reasoning (LNR)}
models the implicit transition from the observed narrative state to its
continuation through compact latent reasoning states. Rather than
autoregressively generating textual rationales, LNR performs internal
narrative inference and directly conditions SID generation, reducing
reasoning overhead while maintaining structured prediction capability.

To facilitate evaluation of this task, we introduce a user-agnostic
multimodal short-form drama continuation benchmark with aligned
story contexts and ground-truth successor segments across UGC, PGC, and
OOD settings. Extensive experiments demonstrate that NarraLite achieves
superior continuation accuracy and a more favorable accuracy--efficiency
trade-off compared with retrieval-based, sequential, multimodal, and
generative recommendation baselines.

Our contributions are summarized as follows:

\begin{itemize}
    \item We formulate efficient multimodal narrative-aware generative
    recommendation, where the goal is to predict a coherent story
    continuation rather than user-preference-driven item relevance, and
    identify two key efficiency challenges caused by redundant visual
    perception and explicit reasoning generation.

    \item We propose NarraLite, which introduces Progressive Spectral
    Compression for retaining compact and text-complementary visual
    evidence, and Latent Narrative Reasoning for modeling implicit story
    transitions without generating explicit reasoning chains.

    \item We establish a user-agnostic multimodal continuation benchmark
    and extensive experiments demonstrate that NarraLite consistently
    improves recommendation accuracy, narrative coherence, and inference
    efficiency.
\end{itemize}

%% file: figures/para.tex
\begin{figure}[t]
  \centering
\includegraphics[width=1.0\linewidth]{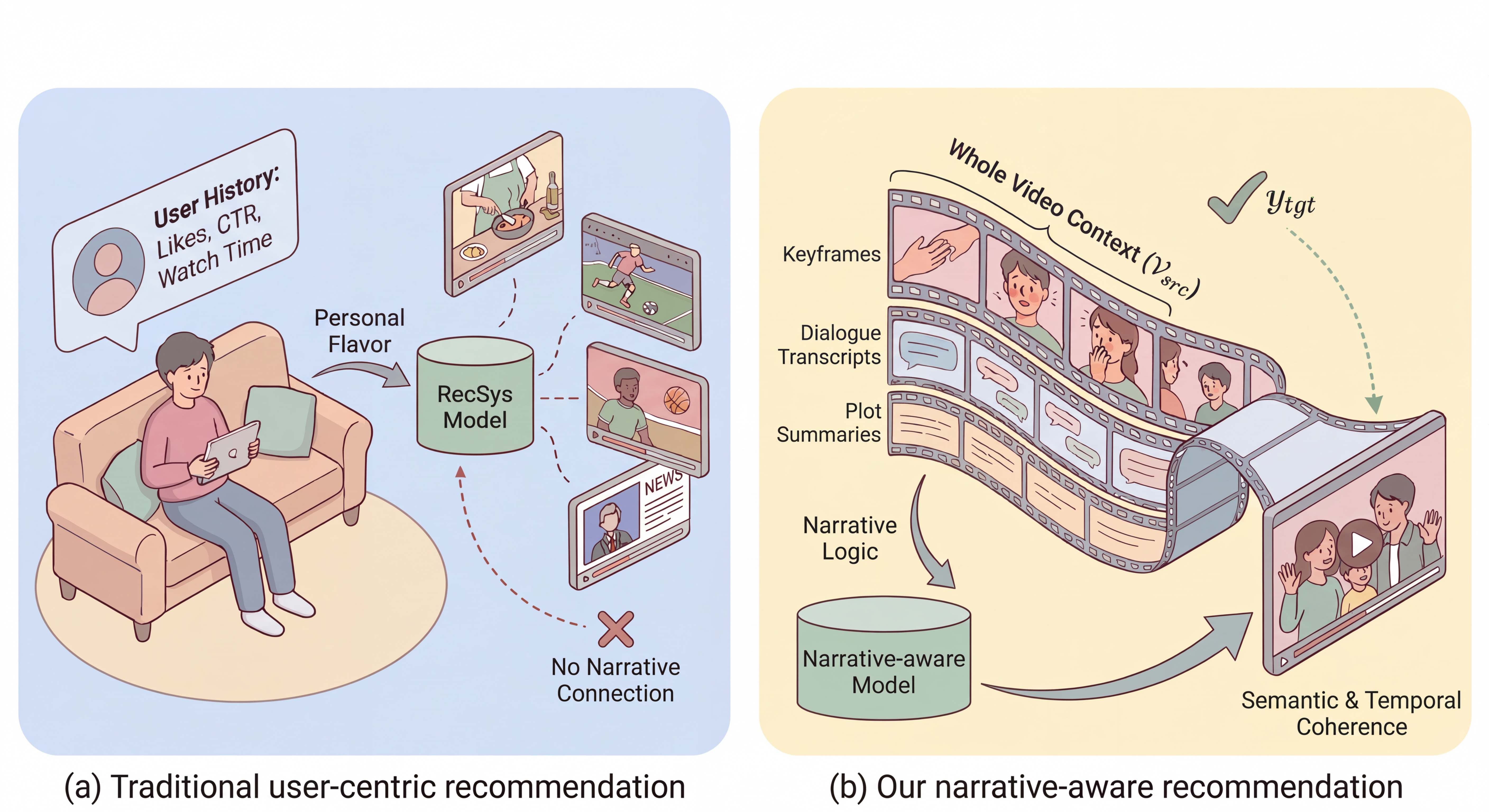}
  \caption{Paradigm shift from user-centric recommendation to narrative-aware recommendation.}
  \label{fig:intro}
\end{figure}

%% file: secs/related.tex
\input{figures/model}
\section{Related Work}
\label{sec:related_work}

\subsection{Multimodal Recommendation}
\label{sec:related_multimodal}

Multimodal recommendation has evolved from feature-level fusion toward
adaptive alignment and foundation-model-based representation learning.
Early studies focus on integrating heterogeneous item modalities:
MTSTRec~\citep{hong2025mtstrec} introduces modality-shared tokens to coordinate temporal
textual, visual, and price information, while FindRec~\citep{wang2025findrec} further improves
multimodal interaction modeling through adaptive routing and efficient
sequence modeling. Recent works
study robustness and efficiency under imperfect multimodal inputs.
MoDiCF~\citep{yang2026structured} and SSR~\citep{li2025generating} respectively explore generative modality recovery and
frequency-aware multimodal denoising to improve representation quality. With the emergence of multimodal LLM, LaViC~\citep{jeon2025adapting} and PRIME~\citep{yue2025preference} further investigate visual token
compression and LMM-based recommendation pipelines. Despite their progress, existing multimodal recommenders mainly target
user preference modeling and item matching. Their multimodal
representations are optimized for identifying relevant items rather
than preserving sparse visual evidence and inferring cross-segment event
transitions.

\subsection{CoT Reasoning for Recommendation}
\label{sec:related_cot}

Recent studies introduce reasoning mechanisms into recommendation by
using large language models to improve preference understanding.
RecZero~\citep{kong2026think} and R2ec~\citep{you2026r} explore explicit reasoning chains with reinforcement
learning or joint reasoning-prediction optimization, demonstrating the
potential of structured reasoning for recommendation. However, explicit CoT introduces
additional autoregressive decoding costs, and recent analyses such as
MME-CoT~\citep{jiang2025mme} reveal that excessive reasoning may bring unnecessary inference
overhead in multimodal scenarios. To improve efficiency, subsequent works attempt to internalize
reasoning into latent representations. SIREN~\citep{dingtoken} distills explicit
preference reasoning into hidden states, while LatentR3~\citep{zhang2025reinforced} and IntuRec~\citep{liu2026intuition}
further explore latent reasoning tokens for direct recommendation
generation.
Nevertheless, existing reasoning-based recommenders primarily model
latent user preferences from interaction histories or candidate items.
They neither ground latent states in multimodal narrative evidence nor
capture the transition from observed events to future story states.

%% file: figures/model.tex
\begin{figure*}
    \centering
    \includegraphics[width=1.0\linewidth]{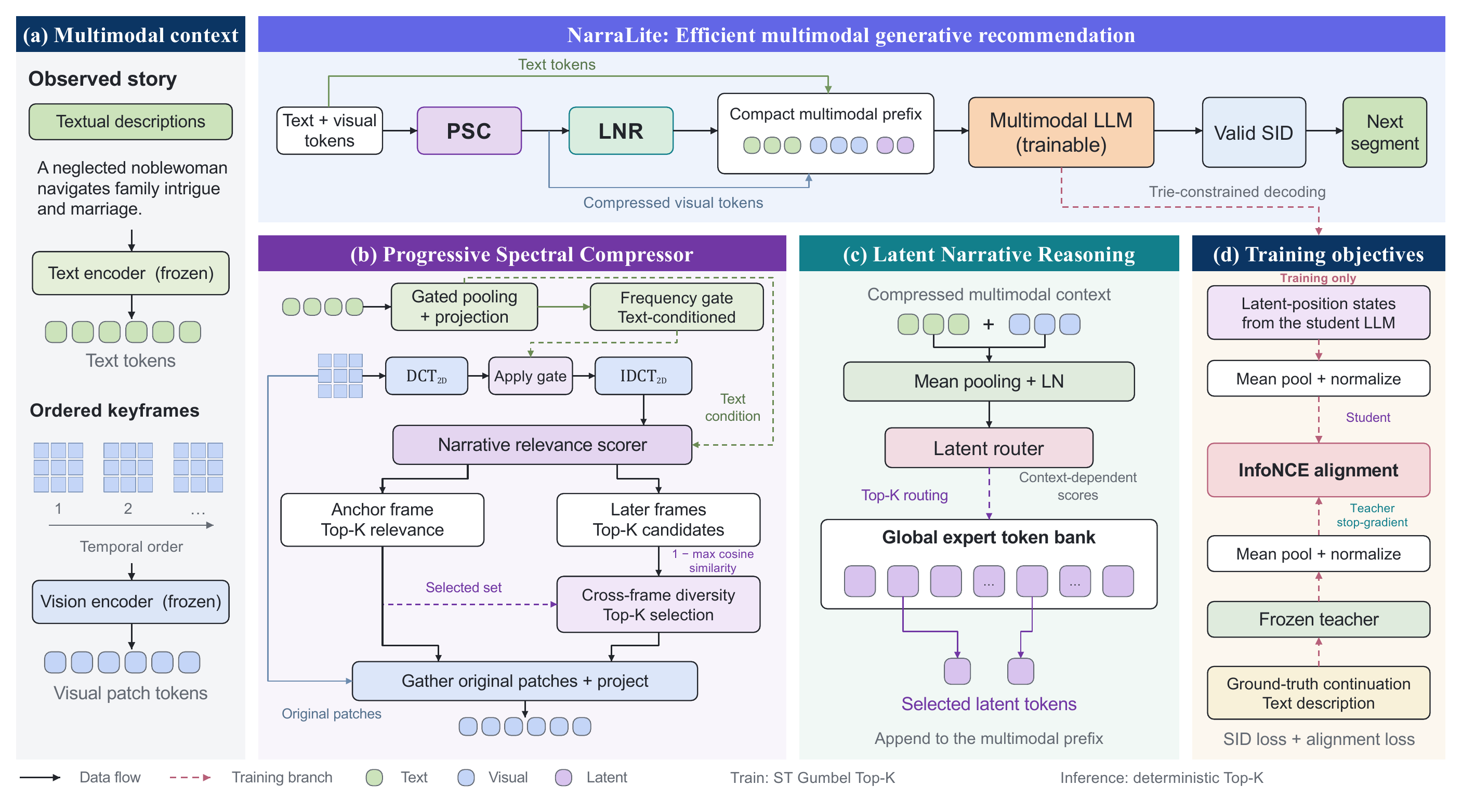}
    \caption{Overview of \textbf{NarraLite}.}
    \label{fig:model}
\end{figure*}

%% file: secs/method.tex
\section{Preliminaries}
\label{sec:preliminaries}

\subsection{Problem Formulation}
\label{sec:problem}

Let $\mathcal{Y}=\{y_j\}_{j=1}^{|\mathcal{Y}|}$ denote the candidate
video segments. We define the training set as narrative continuation
pairs:
\begin{equation}
    \mathcal{D}
    =
    \left\{
    (\mathcal{C}_i,y_i^+)
    \right\}_{i=1}^{N},
    \qquad
    y_i^+\in\mathcal{Y},
\end{equation}
where $\mathcal{C}_i$ denotes the observed narrative context and
$y_i^+$ is the ground-truth continuation. Each context consists of
temporally ordered multimodal segments:
\begin{equation}
    \mathcal{C}_i
    =
    \left\{
    (\mathcal{F}_{i,s},\mathcal{X}_{i,s})
    \right\}_{s=1}^{S_i},
\end{equation}
where $\mathcal{F}_{i,s}$ and $\mathcal{X}_{i,s}$ represent visual
keyframes and textual information, respectively. Different from
conventional recommendation, the target segment is determined by
intrinsic storyline progression rather than user preference.

We formulate continuation prediction as semantic identifier generation.
A fixed multimodal encoder $\phi(\cdot)$ and residual quantizer
$\mathcal{Q}(\cdot)$ encode each candidate segment as:
\begin{equation}
    \operatorname{SID}(y_j)
    =
    \mathcal{Q}
    \left(
    \phi(y_j)
    \right)
    =
    [c_j^1,\ldots,c_j^L].
\end{equation}
The model autoregressively generates the SID of the ground-truth
continuation:
\begin{equation}
    p_{\Theta}
    \left(
    \operatorname{SID}(y_i^+)
    |
    \mathcal{C}_i
    \right)
    =
    \prod_{\ell=1}^{L}
    p_{\Theta}
    \left(
    c_i^\ell
    |
    \mathcal{C}_i,c_i^{<\ell}
    \right).
\end{equation}
During inference, trie-constrained decoding produces valid SIDs, which
are deterministically mapped back to their corresponding segments.

\section{Method}
\label{sec:method}

\subsection{Overview}
\label{sec:overview}

We propose \textbf{NarraLite}, an efficient multimodal generative
recommendation framework for narrative continuation. As illustrated in
Figure~\ref{fig:model}, NarraLite consists of two major components:
\textbf{Progressive Spectral Compressor} (PSC) and
\textbf{Latent Narrative Reasoning} (LNR).

PSC aims to reduce redundant visual computation while preserving
transition-critical visual evidence. Instead of uniformly sampling or
compressing visual tokens through soft aggregation, PSC performs
text-conditioned spectral modulation followed by relevance-aware and
diversity-aware token selection. LNR introduces a mixture-of-expert
latent token bank to capture future narrative tendencies without
explicitly generating textual reasoning chains.

Given the narrative context $\mathcal{C}_i$, we first aggregate all
visual and textual observations:
\begin{equation}
\begin{aligned}
    \mathcal{X}_i
    &=
    \operatorname{Concat}
    (
    \mathcal{X}_{i,1},\ldots,\mathcal{X}_{i,S_i}
    ),
    \\
    \mathcal{F}_i
    &=
    \operatorname{Concat}
    (
    \mathcal{F}_{i,1},\ldots,\mathcal{F}_{i,S_i}
    ).
\end{aligned}
\end{equation}

A frozen text encoder and vision encoder produce:
\begin{equation}
\begin{aligned}
    \mathbf{T}_i
    &=
    f_{\mathrm{txt}}
    (
    \mathcal{X}_i
    )
    \in
    \mathbb{R}^{M_i\times d},
    \\
    \mathbf{V}_i
    &=
    f_{\mathrm{vis}}
    (
    \mathcal{F}_i
    )
    \in
    \mathbb{R}^{N_i\times H\times W\times d_v},
\end{aligned}
\end{equation}
where the visual context contains
$P_i=N_iHW$ patch tokens.

PSC compresses visual features into $K_v$ tokens:
\begin{equation}
    \mathbf{V}_i^c
    =
    \operatorname{PSC}
    (
    \mathbf{V}_i,\mathbf{T}_i
    )
    \in
    \mathbb{R}^{K_v\times d},
    \qquad
    K_v\ll P_i .
\end{equation}

The compressed visual tokens and textual tokens are further processed by
LNR to obtain $K_r$ latent narrative tokens:
\begin{equation}
    \mathbf{R}_i
    =
    \operatorname{LNR}
    (
    \mathbf{T}_i,\mathbf{V}_i^c
    )
    \in
    \mathbb{R}^{K_r\times d}.
\end{equation}

Finally, the multimodal prefix is constructed as:
\begin{equation}
    \mathbf{Z}_i
    =
    [
    \mathbf{T}_i;
    \mathbf{V}_i^c;
    \mathbf{R}_i
    ],
\end{equation}
which is fed into the multimodal LLM for SID generation.

\subsection{Progressive Spectral Compressor}
\label{sec:psc}

Long video contexts contain redundant spatial and temporal observations,
while directly removing tokens may discard subtle but important events.
PSC therefore performs narrative-aware visual compression through three
steps: text-conditioned spectral modulation, patch relevance scoring,
and cross-frame diversity selection.

\subsubsection{Text-Conditioned Spectral Modulation}

We first obtain a narrative condition from textual tokens. For each text
token $\mathbf{t}_{i,m}$:
\begin{equation}
    a_{i,m}
    =
    \sigma
    (
    \mathbf{w}_p^\top\mathbf{t}_{i,m}+b_p
    ),
\end{equation}
and compute gated pooling:
\begin{equation}
    \bar{\mathbf t}_i
    =
    \frac{
    \sum_m a_{i,m}\mathbf{t}_{i,m}
    }
    {
    \sum_m a_{i,m}+\epsilon
    }.
\end{equation}

The textual condition is projected into the visual space:
\begin{equation}
    \mathbf q_i
    =
    \operatorname{LN}
    (
    \mathbf W_t\bar{\mathbf t}_i+\mathbf b_t
    ).
\end{equation}

For each frame, PSC transforms visual patches into frequency space:
\begin{equation}
    \hat{\mathbf V}_{i,n}
    =
    \operatorname{DCT}_{2D}
    (
    \mathbf V_{i,n}
    ).
\end{equation}

A sample-specific frequency gate is generated:
\begin{equation}
    \mathbf G_i
    =
    \sigma
    (
    \operatorname{MLP}_f(\mathbf q_i)
    ),
\end{equation}
and applied before inverse transformation:
\begin{equation}
    \tilde{\mathbf V}_{i,n}
    =
    \operatorname{IDCT}_{2D}
    (
    \mathbf G_i
    \odot
    \hat{\mathbf V}_{i,n}
    ).
\end{equation}

This operation allows textual narrative information to adaptively
preserve either coarse scene structure or fine-grained visual details.

\subsubsection{Narrative-Aware Token Selection}

After spectral modulation, PSC assigns each patch a relevance score:
\begin{equation}
    s_{i,n,p}
    =
    \mathbf w_s^\top
    \operatorname{GELU}
    (
    \mathbf W_s
    (
    \tilde{\mathbf v}_{i,n,p}
    \odot
    \mathbf q_i
    )
    +
    \mathbf b_s
    ).
\end{equation}

For the anchor frame, PSC directly selects the top-$K_a$ patches:
\begin{equation}
    \mathcal I_{i,1}
    =
    \operatorname{STGumbelTopK}
    (
    \mathbf s_{i,1},
    K_a
    ).
\end{equation}

For subsequent frames, PSC first constructs a candidate pool:
\begin{equation}
    \mathcal P_{i,n}
    =
    \operatorname{STGumbelTopK}
    (
    \mathbf s_{i,n},
    K_c
    ),
    \qquad K_c>K_f .
\end{equation}

To avoid redundant selections across frames, we compute diversity scores:
\begin{equation}
    d_{i,n,p}
    =
    1-
    \max_{q\in\mathcal I_i}
    \operatorname{cos}
    (
    \mathbf v_{i,n,p},
    \mathbf v_{i,q}
    ).
\end{equation}

The final tokens are selected according to:
\begin{equation}
    \mathcal I_{i,n}
    =
    \operatorname{STGumbelTopK}
    (
    \{d_{i,n,p}\},
    K_f
    ).
\end{equation}

The selected complete patch embeddings are concatenated and projected:
\begin{equation}
    \mathbf V_i^c
    =
    \operatorname{Proj}_v
    (
    \operatorname{Select}
    (
    \mathbf V_i
    )
    )
    \in
    \mathbb R^{K_v\times d}.
\end{equation}

\subsection{Latent Narrative Reasoning}
\label{sec:lnr}

The compressed multimodal context provides evidence of the observed
storyline, but the model still needs to infer the latent direction toward
the future continuation. Explicit chain-of-thought reasoning introduces
additional autoregressive cost, while directly generating context
dependent latent tokens may simply duplicate information already
captured by the multimodal LLM. 

We therefore introduce a mixture-of-expert latent token bank, where the
semantic space of future narrative tendencies is learned globally and
the router dynamically activates suitable latent tokens for each
instance.

\subsubsection{Context-Routed Expert Token Bank}

We first summarize the observed narrative context:
\begin{equation}
    \mathbf h_i^{ctx}
    =
    \operatorname{LN}
    \left(
    \operatorname{MeanPool}
    (
    [
    \mathbf T_i;
    \mathbf V_i^c
    ]
    )
    \right)
    \in
    \mathbb R^d .
\end{equation}

A lightweight router predicts the relevance of each expert:
\begin{equation}
    \boldsymbol\rho_i
    =
    \mathbf W_r
    \mathbf h_i^{ctx}
    +
    \mathbf b_r
    \in
    \mathbb R^{M_e},
\end{equation}
where $M_e$ is the number of latent experts.

The expert token bank is defined as:
\begin{equation}
    \mathbf E
    =
    [
    \boldsymbol\theta_1;
    \boldsymbol\theta_2;
    \ldots;
    \boldsymbol\theta_{M_e}
    ]
    \in
    \mathbb R^{M_e\times d},
\end{equation}
where each $\boldsymbol\theta_j$ is a learnable continuous token. Unlike
context-dependent latent generation, the expert tokens are independent
of the current input, while the router determines which future narrative
patterns should be activated.

During training, we select $K_r$ experts with straight-through Gumbel
Top-$K$ routing:
\begin{equation}
    \mathbf A_i^r
    =
    \operatorname{STGumbelTopK}
    (
    \boldsymbol\rho_i,K_r
    ),
\end{equation}
and obtain the latent narrative tokens:
\begin{equation}
    \mathbf R_i
    =
    \mathbf A_i^r\mathbf E
    \in
    \mathbb R^{K_r\times d}.
\end{equation}

At inference time, the stochastic routing is replaced with deterministic
Top-$K_r$ selection. The selected latent tokens are appended after the
multimodal observations:
\begin{equation}
    \mathbf Z_i
    =
    [
    \mathbf T_i;
    \mathbf V_i^c;
    \mathbf R_i
    ] .
\end{equation}

The multimodal LLM contextualizes these latent tokens together with the
observed narrative context, allowing them to encode sample-specific
future transition information.

\subsubsection{Target-Guided Latent Alignment}

The SID generation objective alone does not explicitly enforce that the
latent states represent the semantic direction of the future
continuation. We therefore introduce a training-only alignment objective
between latent hidden states and the ground-truth continuation semantics.

During training, the multimodal LLM processes the compact prefix together
with the target SID:
\begin{equation}
    \mathbf H_i^{stu}
    =
    \mathcal M_{\Theta}
    \left(
    [
    \mathbf Z_i;
    \operatorname{Emb}_{\Theta}
    (
    \operatorname{SID}(y_i^+)
    )
    ]
    \right).
\end{equation}

We extract the final hidden states corresponding to the latent positions:
\begin{equation}
    \mathbf H_i^r
    =
    [
    \mathbf h_{i,1}^r;
    \ldots;
    \mathbf h_{i,K_r}^r
    ] .
\end{equation}

The student latent representation is obtained by:
\begin{equation}
    \mathbf z_i^s
    =
    \operatorname{Normalize}
    (
    \operatorname{MeanPool}
    (
    \mathbf H_i^r
    )
    ).
\end{equation}

Meanwhile, a frozen teacher model encodes the textual description of the
ground-truth continuation:
\begin{equation}
    \mathbf H_i^t
    =
    \overline{\mathcal M}
    (
    \mathcal X_i^+
    ),
\end{equation}
and produces:
\begin{equation}
    \mathbf z_i^t
    =
    \operatorname{Normalize}
    (
    \operatorname{stopgrad}
    (
    \operatorname{MeanPool}
    (
    \mathbf H_i^t
    )
    )
    ).
\end{equation}

Given a mini-batch of size $B$, we optimize an InfoNCE objective:
\begin{equation}
    \mathcal L_{\mathrm{NCE}}
    =
    -
    \frac{1}{B}
    \sum_{i=1}^{B}
    \log
    \frac{
    \exp
    (
    (\mathbf z_i^s)^\top
    \mathbf z_i^t/\tau_c
    )
    }
    {
    \sum_{j=1}^{B}
    \exp
    (
    (\mathbf z_i^s)^\top
    \mathbf z_j^t/\tau_c
    )
    },
\end{equation}
where $\tau_c$ is the temperature coefficient.

This alignment encourages latent tokens to capture reusable future
narrative directions without requiring explicit reasoning annotations.

\subsection{Generative Learning and Inference}
\label{sec:learning}

Given the compact multimodal prefix $\mathbf Z_i$, the multimodal LLM
generates the semantic identifier of the ground-truth continuation:
\begin{equation}
    p_{\Theta}
    (
    \operatorname{SID}(y_i^+)
    |
    \mathcal C_i
    )
    =
    \prod_{\ell=1}^{L}
    p_{\Theta}
    (
    c_i^\ell
    |
    \mathbf Z_i,
    c_i^{<\ell}
    ).
\end{equation}

The SID generation loss is:
\begin{equation}
    \mathcal L_{\mathrm{SID}}
    =
    -
    \frac{1}{B}
    \sum_{i=1}^{B}
    \sum_{\ell=1}^{L}
    \log
    p_{\Theta}
    (
    c_i^\ell
    |
    \mathbf Z_i,
    c_i^{<\ell}
    ).
\end{equation}

The overall training objective combines SID generation and latent
narrative alignment:
\begin{equation}
    \mathcal L
    =
    \mathcal L_{\mathrm{SID}}
    +
    \lambda
    \mathcal L_{\mathrm{NCE}},
\end{equation}
where $\lambda$ controls the contribution of latent supervision.

During inference, the teacher branch and alignment objective are removed.
PSC and LNR use deterministic Top-$K$ selection, and the multimodal LLM
generates valid SIDs through trie-constrained decoding:
\begin{equation}
    \widehat{\operatorname{SID}}_i
    =
    \arg\max_{\mathbf c\in\mathcal S_{\mathrm{SID}}}
    p_{\Theta}
    (
    \mathbf c|\mathbf Z_i
    ).
\end{equation}

The predicted identifier is finally mapped back to its corresponding
continuation segment:
\begin{equation}
    \hat y_i
    =
    \operatorname{SID}^{-1}
    (
    \widehat{\operatorname{SID}}_i
    ).
\end{equation}

Compared with processing all visual tokens, NarraLite only introduces
$M_i+K_v+K_r$ prefix tokens instead of $M_i+P_i$. Moreover, latent
tokens are processed in parallel during LLM inference rather than being
autoregressively generated as explicit reasoning chains, reducing both
visual computation and reasoning latency.

%% file: secs/exp.tex
\input{tables/stat}

\section{Experimental Setup}
\label{sec:setup}
\subsection{Dataset}

We evaluate our method on a user-agnostic multimodal narrative
continuation benchmark that we construct for short-form dramas. In this benchmark, each sample contains ordered visual frames, textual descriptions, and
the corresponding continuation segment, enabling evaluation of
narrative-aware recommendation without user preference bias. As shown in Table~\ref{tab:stat}, our dataset comprises over 250,000 video pairs across five splits: Train, Eval, and three diverse test sets designed to evaluate narrative generalization: User-Generated Content (UGC), Professional Generated Content (PGC), and Out-of-Distribution (OOD). The OOD split isolates the concluding episodes of PGC dramas to test model performance on unseen narrative endings. While PGC and OOD segments exhibit a concise, linear progression—averaging 60 seconds with unique successors—the UGC subset introduces significant complexity through higher narrative density (294 works on average) and greater continuation multiplicity (\emph{avg.} 2.4). Details of the benchmark can be found in Appendix A.


\subsection{Metrics}

We evaluate exact continuation retrieval using standard H@K, which
measures whether the ground-truth segment $y_i^{+}$ appears in the top-$K$
generated candidates.

To evaluate whether the model captures intrinsic narrative coherence
rather than relying on superficial visual similarity, we introduce
\textbf{Overlap-Penalized Semantic Coherence (OPSC)}. Given the top-$K$
predicted segments $\{\hat{y}_{i,1},\ldots,\hat{y}_{i,K}\}$ and the
ground-truth continuation $y_i^{+}$, OPSC measures their semantic
alignment while penalizing excessive visual overlap:
\begin{equation}
    \mathrm{OPSC}
    =
    \frac{1}{K}
    \sum_{k=1}^{K}
    \operatorname{sim}
    \left(
    \phi(\hat{y}_{i,k}),
    \phi(y_i^{+})
    \right)
    \cdot
    \exp
    \left(
    -\beta R(\hat{y}_{i,k},y_i^{+})
    \right),
\end{equation}
where $\phi(\cdot)$ denotes the multimodal segment encoder defined in
Section~\ref{sec:problem}, $\operatorname{sim}(\cdot,\cdot)$ measures
semantic similarity, and $R(\hat{y}_{i,k},y_i^{+})$ represents the visual
overlap ratio between the predicted and ground-truth segments. The
hyperparameter $\beta$ controls the strength of the overlap penalty.
Higher OPSC indicates that the predicted continuation is semantically
consistent with the ground truth while avoiding trivial visual
duplication. Together, H@K and OPSC evaluate both exact continuation
accuracy and intrinsic narrative progression.

\input{tables/main}
\input{tables/opsc}

\subsection{Baselines}

We compare NarraLite with representative methods from three categories. \textbf{Retrieval-based methods:} including BM25 and Vector Retrieval, which measure continuation relevance based on textual or multimodal semantic similarity. \textbf{Sequential recommendation models:} including SASRec~\citep{kang2018self}, GRU4Rec~\citep{shehzad2025revisiting}, and BERT4Rec~\citep{sun2019bert4rec}, which model sequential dependencies from historical interactions and are adapted to the narrative continuation setting. \textbf{Generative recommendation models:} including TIGER~\citep{rajput2023recommender}, LETTER~\citep{wang2024learnable}, EARec~\citep{yang2025explainable}, and LatentR3~\citep{zhang2025reinforced}. TIGER and LETTER represent the standard paradigm of generating discrete semantic identifiers for target items, while EARec and LatentR3 further introduces visual and latent reasoning tokens to improve recommendation generation without explicit reasoning chains. The details can be found in Appendix B.

\subsection{Implementation Details}
\label{sec:implementation}

We implement NarraLite based on the pretrained multimodal large language
model Qwen-3.5-2B~\citep{qwen35_blog}. Following prior generative
recommendation frameworks, we construct semantic identifiers using
RQ-KMeans~\citep{xu2025mmq}. Specifically, a three-level residual
quantization scheme with codebook sizes of $(512,256,256)$ is applied to
multimodal segment representations, producing hierarchical SIDs for all
candidate segments.

All experiments are conducted on 8 NVIDIA H20 GPUs. During inference, we employ trie-constrained beam search
with 10 generated candidates for each query. Unless otherwise
specified, NarraLite retains 30 compressed visual tokens
($K_v=30$) and 2 latent narrative tokens ($K_r=2$), which are selected
according to the hyperparameter analysis.

\section{Experimental Results}
\label{sec:res}

\subsection{Main Results}

Table~\ref{tab:main_results} summarizes the overall performance
comparison. We have the following observations.

\textbf{Generative recommendation is superior to retrieval-based
approaches.}
Generative models consistently outperform retrieval methods across all
splits. For example, NarraLite achieves an H@10 of 0.3933 on Eval,
significantly exceeding Dense Retrieval (0.2569), demonstrating that
direct SID generation can capture high-level narrative transitions
beyond surface-level similarity matching. This advantage is further
maintained under OOD scenarios, where retrieval methods suffer from
limited generalization due to their dependence on explicit semantic
overlap.

\textbf{Latent reasoning and multimodal evidence are complementary.}
LatentR3 achieves stronger performance than previous generative
baselines, especially on OOD, validating the effectiveness of latent
reasoning for modeling future narrative states. NarraLite further
improves over LatentR3 by jointly leveraging compressed visual evidence
and latent narrative reasoning, achieving the best performance across
all evaluation settings. These results demonstrate that accurate
continuation requires both multimodal understanding of observed events
and implicit modeling of future transitions.

\textbf{NarraLite achieves the best overall accuracy.}
NarraLite obtains the highest H@1, H@3, H@5, and H@10 on all splits,
achieving an H@10 of 0.3933 on Eval and 0.3090 on OOD. The consistent
improvements verify that Progressive Spectral Compression effectively
preserves narrative-relevant visual information, while Latent Narrative
Reasoning enables efficient prediction of future storyline states
without explicit reasoning chains.

\subsection{Narrative Coherence Analysis}
Table~\ref{tab:opsc_results} reports Overlap-Penalized Semantic
Coherence (OPSC), which evaluates whether predicted segments remain
semantically consistent with the ground-truth continuation after
penalizing excessive visual overlap. Generative recommenders consistently
outperform the multimodal matching baseline, indicating that SID
generation captures narrative relationships beyond static cross-modal
similarity. Among the generative baselines, LatentR3 achieves stronger
coherence, particularly under structured and out-of-distribution
storylines, highlighting the benefit of latent reasoning for modeling
future narrative states. NarraLite further obtains the best OPSC across
all evaluation settings, suggesting that narrative-relevant visual
compression and latent transition reasoning are complementary. Thus,
even when the exact target is not ranked first, NarraLite tends to
produce continuations that remain logically aligned
with the intended storyline rather than relying on superficial visual
repetition.

\input{figures/hyper}

\subsection{Ablation Study}

We conduct ablation studies to investigate the contribution of the two
key components in NarraLite: multimodal perception and latent narrative
reasoning. As shown in Table~\ref{tab:main_results}, both variants
consistently degrade performance across evaluation settings, confirming
the necessity of each component.

\textbf{Effect of latent narrative reasoning.}
Removing the latent reasoning module (w/o LCOT) leads to noticeable
performance drops, particularly on the OOD split. This indicates that
explicitly modeling the transition from observed events to future
narrative states is important for generalizing beyond seen story
patterns. Without latent reasoning, the model mainly relies on direct
context matching, making it harder to infer implicit storyline
progressions.

\textbf{Effect of multimodal perception.}
Removing multimodal inputs (w/o MM) also decreases performance,
especially on UGC scenarios where visual evidence is often essential for
disambiguating characters, actions, and scene changes. This demonstrates
that textual descriptions alone cannot fully capture the visual plot required for narrative continuation.


\subsection{Hyperparameter Analysis}
\label{sec:hyperparameter}

We analyze the sensitivity of NarraLite to the number of retained visual
tokens $K_v$ and latent reasoning tokens $K_r$ on the Eval split, while
keeping all other settings fixed. As shown in
Figure~\ref{fig:hyperparameter}, increasing $K_v$ initially improves the
performance, which peaks at $K_v=30$. Retaining fewer tokens may discard
plot-critical visual evidence, whereas using more tokens reintroduces
redundant information and weakens the benefits of visual compression.
For latent reasoning, the best result is obtained with $K_r=2$. A single
latent token provides limited capacity, while additional tokens may introduce redundant or
conflicting reasoning signals.

\input{figures/effi}

\subsection{Efficiency Analysis}
\label{sec:efficiency}

Figure~\ref{fig:accuracy_throughput} compares the accuracy--throughput
trade-off between NarraLite and full-frame variants using different
input resolutions. Each \textit{Res-$x$} variant retains all visual
patches extracted from frames resized to resolution $x$. Reducing the
resolution generally improves throughput but may remove fine-grained
visual evidence, whereas increasing the resolution introduces
substantially more visual tokens without consistently improving
continuation accuracy. This indicates that global resolution reduction
provides an inefficient trade-off between visual detail and computation.

In contrast, NarraLite lies in the upper-right region of the accuracy--throughput space and strictly dominates all resolution variants. Instead of uniformly degrading the visual input, Progressive Spectral Compression selectively preserves plot-relevant patches while discarding redundant content before multimodal language modeling. These results demonstrate that selective visual-token compression provides a more favorable efficiency--accuracy trade-off than processing complete frames at either low or high resolutions.

%% file: tables/stat.tex
\begin{table}[htbp]
\centering
\caption{The overall statistics of our dataset.}
\vspace{-8pt}
\label{tab:stat}
\resizebox{1.0\linewidth}{!}{
\begin{tabular}{
    l
    |ccccc
}
\toprule
\textbf{Split} & \textbf{Train} & \textbf{Eval} & \textbf{UGC} & \textbf{PGC} & \textbf{OOD} \\
\midrule
\textbf{Pairs} & 241,545 & 14,231 & 11,175 & 1,613 & 1,443 \\
\textbf{Texts} & 268 & 247 & 294 & 76 & 78 \\
\textbf{Duration} & 474 & 431 & 533 & 58 & 60 \\
\textbf{Segments} & 8.7 & 8.1 & 9.6 & 2.6 & 2.7 \\
\textbf{Continuations} & 2.2 & 2.1 & 2.4 & 1 & 1 \\
\bottomrule
\end{tabular}
}
\end{table}

%% file: tables/main.tex
\begin{table*}[t]
\centering
\vspace{-10pt}
\caption{
Performance comparison on narrative-aware recommendation. The best and second-best results among non-ablation methods are highlighted in bold and underline, respectively.
}

\label{tab:main_results}

\setlength{\tabcolsep}{2.6pt}
\renewcommand{\arraystretch}{1.15}

\resizebox{\textwidth}{!}{
\begin{tabular}{lcccc|cccc|cccc|cccc}

\toprule

\multirow{2}{*}{\textbf{Method}}
&
\multicolumn{4}{c|}{\textbf{Eval}}
&
\multicolumn{4}{c|}{\textbf{UGC}}
&
\multicolumn{4}{c|}{\textbf{PGC}}
&
\multicolumn{4}{c}{\textbf{OOD}}
\\

\cmidrule(lr){2-5}
\cmidrule(lr){6-9}
\cmidrule(lr){10-13}
\cmidrule(lr){14-17}

&
H@1&H@3&H@5&H@10
&
H@1&H@3&H@5&H@10
&
H@1&H@3&H@5&H@10
&
H@1&H@3&H@5&H@10
\\

\midrule

\methodgroup{Retrieval}

BM25
&0.0715&0.1285&0.1585&0.2022
&0.0827&0.1429&0.1754&0.2241
&0.0254&0.0468&0.0561&0.0827
&0.0173&0.0335&0.0462&0.0618
\\

Dense
&0.0860&0.1655&0.2053&0.2569
&0.0879&0.1694&0.2107&0.2638
&0.0704&0.1287&0.1621&0.2186
&0.0788&0.1475&0.1812&0.2127
\\

\midrule

\methodgroup{Sequential}

SASRec
&0.0595&0.0984&0.1153&0.1382
&0.0637&0.1041&0.1226&0.1472
&0.0508&0.0869&0.1017&0.1261
&0.0419&0.0718&0.0874&0.1085
\\

GRU4Rec
&0.0538&0.0874&0.1064&0.1312
&0.0559&0.0918&0.1105&0.1351
&0.0459&0.0754&0.0916&0.1167
&0.0378&0.0662&0.0801&0.1024
\\

BERT4Rec
&0.0570&0.0987&0.1189&0.1405
&0.0613&0.1016&0.1221&0.1463
&0.0498&0.0847&0.1036&0.1288
&0.0426&0.0729&0.0879&0.1079
\\

\midrule

\methodgroup{Multimodal}

EARec
&$\underline{0.1963}$
&0.2760
&0.3103
&0.3483
&$\underline{0.2084}$
&0.2898
&0.3275
&0.3681
&0.1715
&0.2449
&0.2804
&0.3145
&0.1368
&0.1938
&0.2257
&0.2596
\\

\midrule

\methodgroup{Generative}

TIGER
&0.1888&0.2768&0.3108&0.3458
&0.2019&0.2945&0.3338&0.3757
&0.1654&0.2408&0.2795&0.3140
&0.1286&0.1901&0.2244&0.2578
\\

LETTER
&0.1798&0.2689&0.3032&0.3379
&0.1917&0.2817&0.3205&0.3619
&0.1576&0.2314&0.2687&0.3019
&0.1208&0.1815&0.2148&0.2487
\\

LatentR3
&0.1908
&$\underline{0.2773}$
&$\underline{0.3155}$
&$\underline{0.3583}$
&0.2031
&$\underline{0.3018}$
&$\underline{0.3407}$
&$\underline{0.3785}$
&$\underline{0.1752}$
&$\underline{0.2569}$
&$\underline{0.2975}$
&$\underline{0.3318}$
&$\underline{0.1497}$
&$\underline{0.2186}$
&$\underline{0.2574}$
&$\underline{0.2910}$
\\

\midrule

\oursgroup{Ours}

NarraLite
&$\mathbf{0.1988}$
&$\mathbf{0.3068}$
&$\mathbf{0.3480}$
&$\mathbf{0.3933}$
&$\mathbf{0.2175}$
&$\mathbf{0.3352}$
&$\mathbf{0.3810}$
&$\mathbf{0.4218}$
&$\mathbf{0.1861}$
&$\mathbf{0.2735}$
&$\mathbf{0.3192}$
&$\mathbf{0.3570}$
&$\mathbf{0.1598}$
&$\mathbf{0.2328}$
&$\mathbf{0.2724}$
&$\mathbf{0.3090}$
\\

\midrule

\ablationgroup{Ablation}

w/o LNR
&0.2028&0.2915&0.3245&0.3625
&0.2148&0.3198&0.3582&0.3950
&0.1751&0.2540&0.2935&0.3250
&0.1408&0.2036&0.2398&0.2700
\\

w/o PSC
&0.1985&0.2963&0.3373&0.3783
&0.2042&0.3057&0.3458&0.3820
&0.1787&0.2615&0.3004&0.3350
&0.1531&0.2207&0.2591&0.2980
\\

\bottomrule

\end{tabular}
}
\vspace{-10pt}
\end{table*}

%% file: tables/opsc.tex

\definecolor{groupblue}{RGB}{235,240,248}
\definecolor{oursgreen}{RGB}{232,245,233}

\newcommand{\opscgroup}[1]{%
    \addlinespace[4pt]
    \rowcolor{groupblue}
    \textcolor{black!70}{\textbf{\textit{#1}}}
    & & & & \\
    \addlinespace[1pt]
}

\newcommand{\opscoursgroup}[1]{%
    \addlinespace[4pt]
    \rowcolor{oursgreen}
    \textcolor{black!80}{\textbf{\textit{#1}}}
    & & & & \\
    \addlinespace[1pt]
}

\begin{table}[t]
\centering

\caption{
Overlap-Penalized Semantic Coherence (OPSC) results on narrative-aware recommendation.
}

\label{tab:opsc_results}

\setlength{\tabcolsep}{13pt}
\renewcommand{\arraystretch}{1.15}

\resizebox{0.5\textwidth}{!}{
\begin{tabular}{l c|c|c|c}

\toprule

\textbf{Method}
&
\textbf{Eval}
&
\textbf{UGC}
&
\textbf{PGC}
&
\textbf{OOD}
\\

\midrule

\opscgroup{Multimodal}

EARec
& 0.4392
& 0.4817
& 0.3716
& 0.3489
\\

\midrule

\opscgroup{Generative}

TIGER
& 0.5286
& 0.5954
& 0.4768
& 0.4512
\\

LETTER
& 0.5143
& 0.5819
& 0.4597
& 0.4338
\\

LatentR3
& $\underline{0.5587}$
& $\underline{0.6078}$
& $\underline{0.5162}$
& $\underline{0.5051}$
\\

\midrule

\opscoursgroup{Ours}

NarraLite
& $\mathbf{0.5904}$
& $\mathbf{0.6367}$
& $\mathbf{0.5579}$
& $\mathbf{0.5506}$
\\

\bottomrule

\end{tabular}
}

\end{table}

%% file: figures/hyper.tex
\begin{figure}[t]
    \centering
      \begin{subfigure}{0.48\linewidth}
        \includegraphics[width=\linewidth]{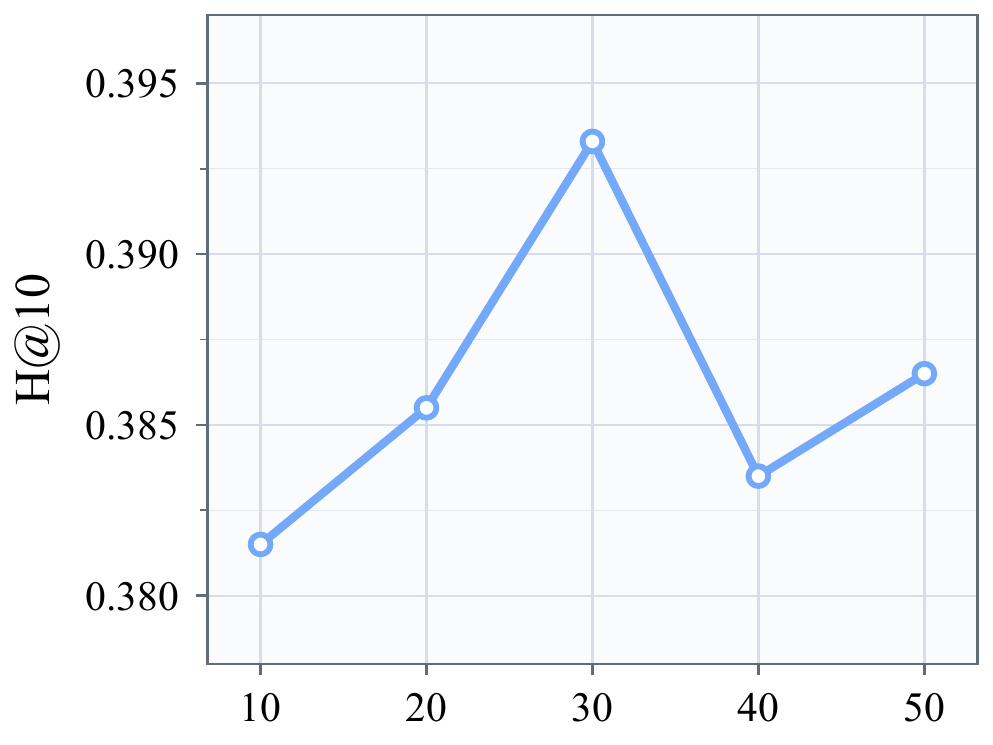}
        \caption{Visual tokens}
        \label{exp:vs}
      \end{subfigure}%
      \hspace{0.5mm}
      \begin{subfigure}{0.48\linewidth}
        \includegraphics[width=\linewidth]{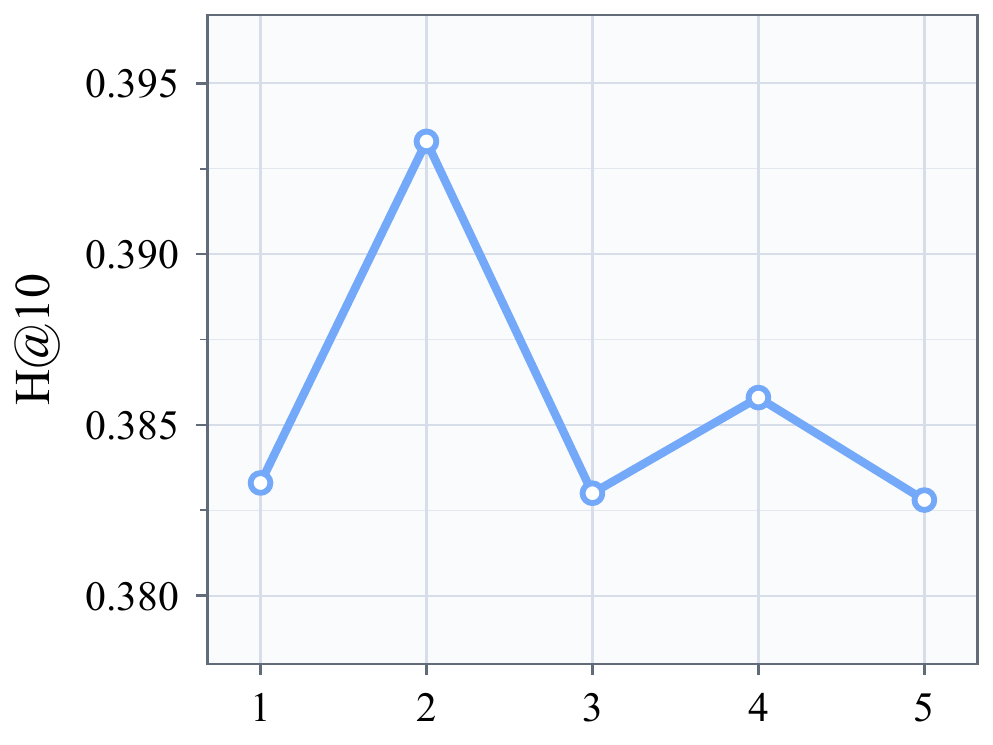}
        \caption{Latent tokens.}
        \label{exp:gr}
      \end{subfigure}%
      \caption{Hyperparameter study.}
      \label{fig:hyperparameter}
\end{figure}

%% file: figures/effi.tex
\begin{figure}[t]
  \centering
\includegraphics[width=1.0\linewidth]{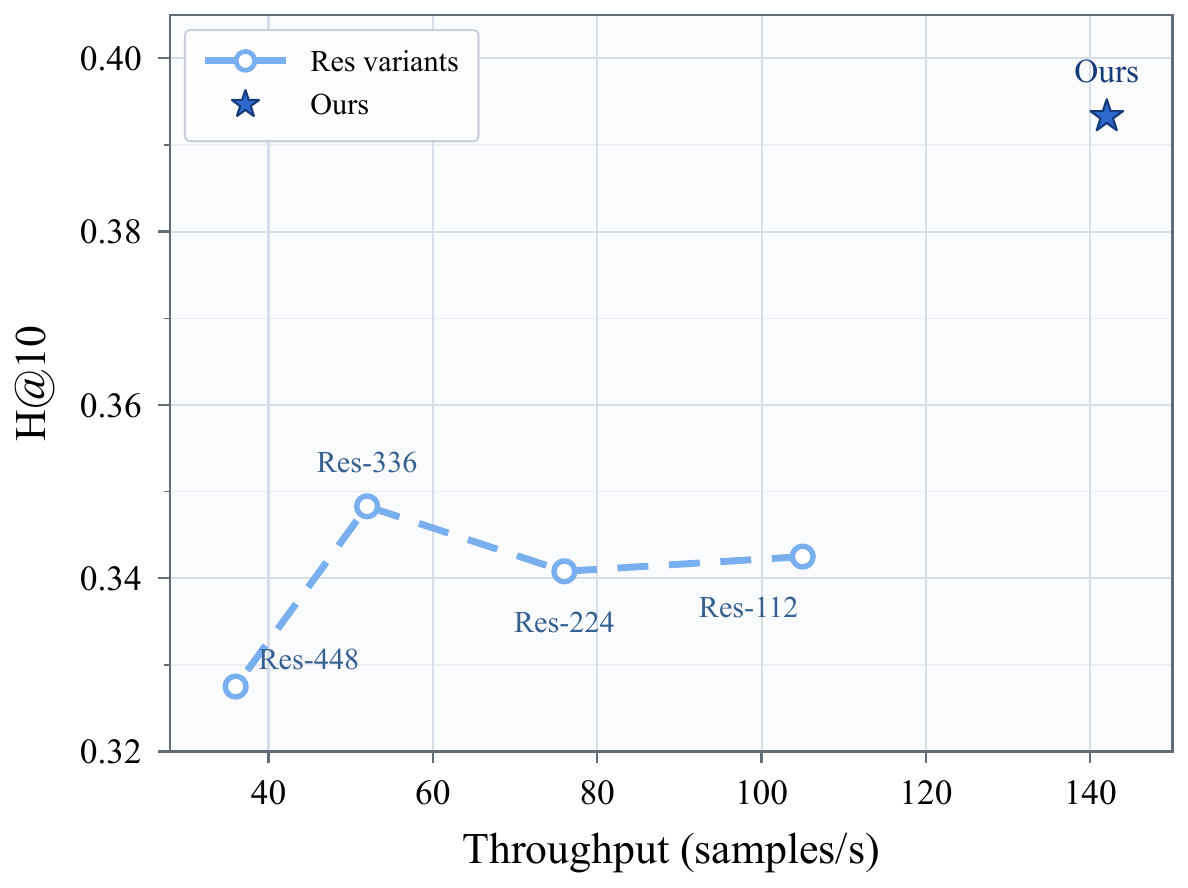}
  \caption{Accuracy--throughput trade-off on the Eval split.}
  \label{fig:accuracy_throughput}
\end{figure}

%% file: secs/conclu.tex
\section{Conclusion}
\label{sec:conclusion}

We studied efficient narrative-aware generative recommendation for
episodic content and proposed \textbf{NarraLite}. Progressive Spectral
Compression selectively preserves compact, text-complementary visual
evidence, while Latent Narrative Reasoning models future storyline
transitions through context-routed expert tokens without explicitly
decoding reasoning chains. Together with a user-agnostic multimodal
benchmark, our experiments demonstrate consistent improvements in
continuation accuracy, out-of-distribution robustness, and inference
efficiency. These results highlight the complementary importance of
selective multimodal perception and latent transition reasoning for
accurate and efficient episodic recommendation.

%% file: aaai2027.bib
@inproceedings{hong2025mtstrec,
  title={MTSTRec: multimodal time-aligned shared token recommender},
  author={Hong, Ming-Yi and Hsu, Yen-Jung and Chiang, Miao-Chen and Lin, Che},
  booktitle={Proceedings of ICML},
  year={2025}
}

@inproceedings{wang2025findrec,
  title={FindRec: Stein-Guided Entropic Flow for Multi-Modal Sequential Recommendation},
  author={Wang, Maolin and Xiao, Yutian and Wang, Binhao and Zhang, Sheng and Ye, Shanshan and Wang, Wanyu and Yin, Hongzhi and Guo, Ruocheng and Xu, Zenglin},
  booktitle={Proceedings of SIGKDD},
  pages={3008--3018},
  year={2025}
}

@inproceedings{yang2026structured,
  title={Structured Spectral Reasoning for Frequency-Adaptive Multimodal Recommendation},
  author={Yang, Wei and Zhong, Rui and Chen, Yiqun and Lu, Chi and Jiang, Peng},
  booktitle={Proceedings of NeurIPS},
  pages={28122--28143},
  year={2026}
}

@inproceedings{li2025generating,
  title={Generating with fairness: A modality-diffused counterfactual framework for incomplete multimodal recommendations},
  author={Li, Jin and Wang, Shoujin and Zhang, Qi and Yu, Shui and Chen, Fang},
  booktitle={Proceedings of WWW},
  pages={2787--2798},
  year={2025}
}

@inproceedings{jeon2025adapting,
  title={Adapting Large Vision-Language Models to Visually-Aware Conversational Recommendation},
  author={Jeon, Hyunsik and Koide, Satoshi and Wang, Yu and He, Zhankui and McAuley, Julian},
  booktitle={Proceedings of SIGKDD},
  pages={1037--1048},
  year={2025}
}

@inproceedings{yue2025preference,
  title={Preference-Optimized Retrieval and Ranking for Efficient Multimodal Recommendation},
  author={Yue, Zhenrui and Zeng, Huimin and Wang, Yueqi and McAuley, Julian and Wang, Dong},
  booktitle={Proceedings of SIGKDD},
  pages={3692--3703},
  year={2025}
}

@inproceedings{kong2026think,
  title={Think before Recommendation: Autonomous Reasoning-enhanced Recommender},
  author={Kong, Xiaoyu and Jiang, Junguang and Liu, Bin and Xu, Ziru and Zhu, Han and Xu, Jian and Zheng, Bo and Wu, Jiancan and Wang, Xiang},
  booktitle={Proceedings of NeurIPS},
  pages={141209--141232},
  year={2026}
}

@inproceedings{you2026r,
  title={R2ec: Towards Large Recommender Models with Reasoning},
  author={You, Runyang and Li, Yongqi and Lin, Xinyu and Zhang, Xin and Wang, Wenjie and Li, Wenjie and Nie, Liqiang},
  booktitle={Proceedings of NeurIPS},
  pages={62376--62405},
  year={2026}
}

@inproceedings{jiang2025mme,
  title={MME-CoT: Benchmarking Chain-of-Thought in Large Multimodal Models for Reasoning Quality, Robustness, and Efficiency},
  author={Jiang, Dongzhi and Zhang, Renrui and Guo, Ziyu and Li, Yanwei and Qi, Yu and Chen, Xinyan and Wang, Liuhui and Jin, Jianhan and Guo, Claire and Yan, Shen and others},
  booktitle={Proceedings of ICML},
  pages={27793--27830},
  year={2025},
}

@inproceedings{dingtoken,
  title={Token-Efficient Long-Term Interest Sketching and Internalized Reasoning for LLM-based Recommendation},
  author={Ding, Zhihao and Li, Jinming and Mu, Shuai and Shi, Jieming},
  booktitle={Proceedings of ICLR},
  year={2026}
}

@article{zhang2025reinforced,
  title={Reinforced latent reasoning for llm-based recommendation},
  author={Zhang, Yang and Xu, Wenxin and Zhao, Xiaoyan and Wang, Wenjie and Feng, Fuli and He, Xiangnan and Chua, Tat-Seng},
  journal={arXiv preprint arXiv:2505.19092},
  year={2025}
}

@article{liu2026intuition,
  title={Intuition-Guided Latent Reasoning for LLM-Based Recommendation},
  author={Liu, Chang and Bai, Yimeng and Zhao, Xiaoyan and Zhang, Yang and Wang, Qifan and Feng, Fuli and Rong, Wenge},
  journal={arXiv preprint arXiv:2606.27684},
  year={2026}
}

@inproceedings{wang2024learnable,
  title={Learnable item tokenization for generative recommendation},
  author={Wang, Wenjie and Bao, Honghui and Lin, Xinyu and Zhang, Jizhi and Li, Yongqi and Feng, Fuli and Ng, See-Kiong and Chua, Tat-Seng},
  booktitle={Proceedings of CIKM},
  pages={2400--2409},
  year={2024}
}

@inproceedings{li2024large,
  title={Large language models for generative recommendation: A survey and visionary discussions},
  author={Li, Lei and Zhang, Yongfeng and Liu, Dugang and Chen, Li},
  booktitle={Proceedings of LREC-COLING},
  pages={10146--10159},
  year={2024}
}

@article{deng2025onerec,
  title={Onerec: Unifying retrieve and rank with generative recommender and iterative preference alignment},
  author={Deng, Jiaxin and Wang, Shiyao and Cai, Kuo and Ren, Lejian and Hu, Qigen and Ding, Weifeng and Luo, Qiang and Zhou, Guorui},
  journal={arXiv preprint arXiv:2502.18965},
  year={2025}
}

@article{chen2025onesearch,
  title={Onesearch: A preliminary exploration of the unified end-to-end generative framework for e-commerce search},
  author={Chen, Ben and Guo, Xian and Wang, Siyuan and Liang, Zihan and Lv, Yue and Ma, Yufei and Xiao, Xinlong and Xue, Bowen and Zhang, Xuxin and Yang, Ying and others},
  journal={arXiv preprint arXiv:2509.03236},
  year={2025}
}

@inproceedings{yue2025cot4rec,
  title={Cot4rec: Revealing user preferences through chain of thought for recommender systems},
  author={Yue, Weiqi and Yin, Yuyu and Zhang, Xin and Shi, Binbin and Liang, Tingting and Wan, Jian},
  booktitle={Proceedings of AAAI},
  pages={13142--13151},
  year={2025}
}

@inproceedings{yang2025explainable,
  title={Explainable Multi-Modality Alignment for Transferable Recommendation},
  author={Yang, Shenghao and Ma, Weizhi and Guo, Zhiqiang and Zhang, Min and Wu, Haiyang and Zhai, Junjie and Zhang, Chunhui and Yang, Yuekui},
  booktitle={Proceedings of WWW},
  pages={2076--2084},
  year={2025}
}

@inproceedings{rajput2023recommender,
  title={Recommender systems with generative retrieval},
  author={Rajput, Shashank and Mehta, Nikhil and Singh, Anima and Keshavan, Raghunandan Hulikal and Vu, Trung and Heldt, Lukasz and Hong, Lichan and Tay, Yi and Tran, Vinh Q and Samost, Jonah and others},
  booktitle={Proceedings of NeurIPS},
  year={2023}
}

@inproceedings{kang2018self,
  title={Self-attentive sequential recommendation},
  author={Kang, Wang-Cheng and McAuley, Julian},
  booktitle={Proceedings of ICDM},
  pages={197--206},
  year={2018},
}

@inproceedings{sun2019bert4rec,
  title={BERT4Rec: Sequential recommendation with bidirectional encoder representations from transformer},
  author={Sun, Fei and Liu, Jun and Wu, Jian and Pei, Changhua and Lin, Xiao and Ou, Wenwu and Jiang, Peng},
  booktitle={Proceedings of CIKM},
  pages={1441--1450},
  year={2019}
}

@inproceedings{shehzad2025revisiting,
  title={Revisiting the Performance of Graph Neural Networks for Session-based Recommendation},
  author={Shehzad, Faisal and Jannach, Dietmar},
  booktitle={Proceedings of RecSys},
  pages={842--846},
  year={2025}
}

@misc{qwen35_blog,
  title        = {Qwen3.5: Towards Native Multimodal Agents},
  author       = {{Qwen Team}},
  year         = {2026},
  howpublished = {\url{https://qwen.ai/blog?id=qwen3.5}},
  note         = {Accessed: 2026-02-16}
}

@article{xu2025mmq,
  title={Mmq: Multimodal mixture-of-quantization tokenization for semantic id generation and user behavioral adaptation},
  author={Xu, Yi and Zhang, Moyu and Li, Chenxuan and Liao, Zhihao and Xing, Haibo and Deng, Hao and Hu, Jinxin and Zhang, Yu and Zeng, Xiaoyi and Zhang, Jing},
  journal={arXiv preprint arXiv:2508.15281},
  year={2025}
}

@inproceedings{zhang2025beyond,
  title={Beyond text-visual attention: Exploiting visual cues for effective token pruning in vlms},
  author={Zhang, Qizhe and Cheng, Aosong and Lu, Ming and Zhang, Renrui and Zhuo, Zhiyong and Cao, Jiajun and Guo, Shaobo and She, Qi and Zhang, Shanghang},
  booktitle={Proceedings of ICCV},
  pages={20857--20867},
  year={2025}
}

@inproceedings{cao2026fastdrivevla,
  title={Fastdrivevla: Efficient end-to-end driving via plug-and-play reconstruction-based token pruning},
  author={Cao, Jiajun and Zhang, Qizhe and Jia, Peidong and Zhao, Xuhui and Lan, Bo and Zhang, Xiaoan and Wei, Xiaobao and Chen, Sixiang and Li, Liyun and Liu, Xianming and others},
  booktitle={Proceedings of AAAI},
  pages={2571--2579},
  year={2026}
}

@inproceedings{he2024multi,
  title={Multi-modal latent space learning for chain-of-thought reasoning in language models},
  author={He, Liqi and Li, Zuchao and Cai, Xiantao and Wang, Ping},
  booktitle={Proceedings of AAAI},
  pages={18180--18187},
  year={2024}
}

@inproceedings{zhan2026l2v,
  title={L2v-cot: Cross-modal transfer of chain-of-thought reasoning via latent intervention},
  author={Zhan, Yu-Liang and Tang, Xinyu and Wan, Han and Li, Jian and Wen, Jirong and Sun, Hao},
  booktitle={Proceedings of AAAI},
  pages={12358--12366},
  year={2026}
}
